# Optimizing Robotic Swarm Based Construction Tasks

Teshan Liyanage
*Universiy of Moratuwa*
Colombo, Sri Lanka
teshanuka@gmail.com

Subha Fernando
*University of Moratuwa*
Colombo, Sri Lanka
subhaf@uom.lk

*Abstract*—Social insects in nature such as ants, termites and bees construct their colonies collaboratively in a very efficient process. In these swarms, each insect contributes to the construction task individually showing redundant and parallel behavior of individual entities. But the robotics adaptations of these swarm's behaviors haven't yet made it to the real world at a large enough scale of commonly being used due to the limitations in the existing approaches to the swarm robotics construction. This paper presents an approach that combines the existing swarm construction approaches which results in a swarm robotic system, capable of constructing a given 2 dimensional shape in an optimized manner.

## I. Introduction

With the emergence of new technologies, construction tasks have become more advanced and less human involved. Meanwhile, swarm robotics has come a long way in the recent past. From transferring social insect behavior to algorithms [1], [2], to plans for using swarm robots for asteroid mining [3], there are many researches done in this field covering a wide variety of applications of swarm robotics. Still the concept has not yet made its way into industrial environments to be used in industrial robots [4].

The advantage of a system of cooperative swarm robots is that they operate locally with minimal resources and without need to understand the complexity of the whole system. Such decentralization of work provides a great robustness and flexibility as a failure of a single entity cannot cause overall system or task failure [5].

Many researchers in the field of construction using swarm robots have come up with successful approaches for constructing simple shapes. However many of them lack practicality due to having features/ capabilities that are not possible to be implemented with existing technologies in the real world. And some do not show crucial behavioral attributes that are expected from a swarm as we will discuss in the next chapter in this paper.

## II. Related Work

Swarm robotic systems adapt social animal behavior to execute tasks collectively [2], [6]-[8]. The behaviors adapted into the system and the usage of the adaptations decides the outcome of such systems. A very simple approach like blindly bulldozing material away by swarm robots can produce simple nests like structures [9]. A more sophisticated, leader based approach can have a leader entity overseeing the construction done by the swarm robots enabling them to construct more complex shapes [10], [11]. A rule based robot controller for swarm robots can produce coherent biological like architectures [1], [12].

A common attribute to many swarm implementations is stigmergic operations. Swarms observed in nature coordinate their behavior stigmergically. Communication and coordination between entities of the swarm happens through the changes in the local environment they observe [12]. This allows robots in a swarm to gain information without heavy communications between robots improving the redundant nature of swarm entities

Most observable and common problem that is present in the existing approaches is the lack of true parallel behavior opposed to what we see in nature. Many swarm approaches make the swarm entities move in parallel. However, the most critical step, which is placing building blocks, happens serially in most cases. Bots depend on the status of the previous block placement in order for the next block placement. Furthermore, in most cases, the construction always happens around one area of the shape at a given time, crowding one area of the construction which increase the construction time because of the increased number of inter robot interactions

In a swarm system, redundancy is a major expectation since there are multiple entities working at the same time. Failure of one entity should not affect the collective task and should not cause the task to fail. Some of the existing approaches show a non-redundant swarm system [13].

In this research we aim at constructing given shapes. In that sense, some of the existing approaches are not successful since they do not aim at constructing a particular shape. The constructed shape emerges from the collective behavior governed by the specified rules for controlling entities in those systems [1].

## III. Approach

Our end goal is to develop a swarm robot system that can be practically used in the industrial construction tasks while overcoming the limitation of the existing approaches. To this end, we hypothesize a swarm of low cost small robots that is practical to be implemented with the existing technologies. The swarm robots are simulated in a 2D graphical environment where the construction tasks can be observed. We input the shape we need constructed to the simulation and the swarm bots construct the given shape using virtual building blocks.

Our approach uses a combination of the existing approaches. We have used concepts of stigmergy and rule based control for the swarm control and have developed a control algorithm (**Algorithm 1**) for the robots to perform their individual tasks so that collectively the shape given is constructed showing expected behaviors.

The most critical task for a successful construction is the systematic order of block placing in the shape for construction. We have designed out building block placement method such that the robots try to build the shape from its center to the outside. Spreading out from the middle of the construction allows robots to work on lesser and lesser crowded areas as the construction progresses, which minimize robot interactions further allowing a faster construction.

### A. Simulation Overview

The simulation environment has swarm robots, building blocks and building block factories. The swarm bots simulated have limited sensory capabilities only to perceive their immediate environment. Bots have a self-localization capability such as GPS, and the ability to identify other bots and building blocks placed. They are not able to go over building blocks or other bots. Bots start with the knowledge of the building block factories and the shape of construction needed to be done. They can carry one building block at a time and need to fetch another one from a building block factory after one is placed.

### B. Placing Building Blocks

The most crucial part of the swarm robot control algorithm is the calculation which determines the order of the building block placement by a robot (Steps 11, 15 and 18 in **Algorithm 1**). Our approach makes a swarm robot start from the centroid of the given shape and construct the shape to the outwards. In this process, the block placement of each robot is biased such that it prioritize the previously placed block's neighborhood area which reduces the probability of all swarm bots clustering near the same location.

The block placement decision making procedure developed can be broken down to the following steps.

*1) Set next block position to the centroid of the given shape*

*2) When placing a block, observe the surrounding blocks and remember the free pixels to be used as next block position candidates :* This encourages bots to stay in one area since the next block position is most probably picked from nearby the area where this block is placed. However, this position might be filled by another bot when the bot returns back with the next block

*3) If the centroid pixel is occupied in the local map, find the closest pixels to the occupied pixels in the map :* Now that there are one or more blocks in the shape, the next block must be placed near the existing blocks. We take the empty pixels that have one or more neighboring occupied pixels (**Fig 1**)

*4) If any of the candidate pixels are in the previously stored candidates, drop the others. Otherwise (if all previously marked candidate pixels are now occupied) do nothing*

*5) If all the candidate pixels have the same number of neighboring occupants, select the one closest as the next block position :* Navigation time is lower for closer blocks

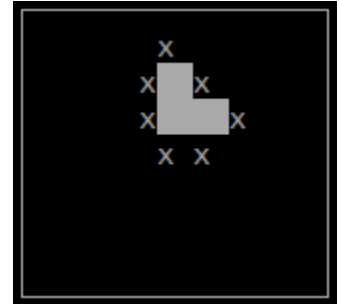

Fig. 1. If white pixels are occupied, we need to place the next blocks at the locations marked by `x`s.

*6) If different candidates have different number of neighboring occupants, select the ones with the highest number of occupant neighbors and select the closest one from among them (**Fig 2**)*

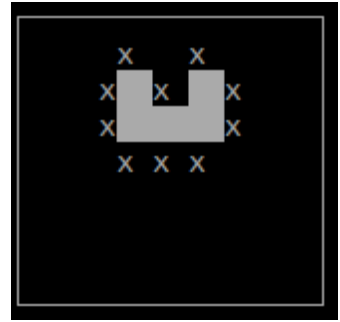

Fig. 2. The pixel stuck inside the u shape has 3 occupant neighbors while others have one or two each. Therefore priority must be given to the one that is closer to be getting trapped inside the shape.

### C. Rules and the Control Algorithm

The rules are introduced to the swarm robots so that they will,

- Depend minimally on each other
- Not obstruct other bots' tasks (no collisions or making bots stuck enclosed in the construction)
- Update their tasks repeatedly based on the new information from the environment

These simple rules are the basis of the control algorithm (**Algorithm 1**)

*1) If travelling and another robot is crossing the travel path, reroute the path considering the other robot as an obstacle*

*2) If the target location to drop the building block is filled, find a new suitable place to drop the building block and reroute to there*

*3) If there is another robot in the vicinity and the next block that is going to be dropped blocks the other's path and make it stuck enclosed inside the construction, find a new suitable place to drop the building block and reroute*

*4) If placing a block makes an empty space in the construction which is not reachable after placing the block, place the block in that empty space instead of the planned location*

```
Algorithm 1: Control algorithm
1  while not all bots finished do
2      foreach bot ∈swarm_bots do
3          Update local_map with sensor data;
4          if stuck or build_shape finished then
5              Set bot finished;
6          else if no block then
7              Set path to nearest block factory;
8              Navigate;
9          else if block factory reached then
10             Load block;
11             Calculate next drop position;
12             Reroute;
13         else if drop position reached then
14             if drop position is filled then
15                 Calculate next drop position;
16                 Reroute;
17             else if drop blocks another bot then
18                 Calculate next drop position;
19                 Reroute;
20             else if drop creates empty unreachable
                   position then
21                 Reroute to unreachable position;
22             else
23                 Drop block;
24                 Reroute to block factory;
25             end
26         else if obstacle on path then
27             Reroute;
28     end
29 end
```

## IV. RESULTS

The aim of the research was to propose a system for an optimized robot swarm. However, for a direct quantitative comparison, very similar researches are needed and no such researches has been found. Each approach proposed in each research tries out a different set of concepts of swarm robotics and tries to improve/ research a different aspect of the field. These vary within self-assembly, 3D construction, wall/ nest construction and beyond.

Considering the above, a qualitative comparison is done and the proposed approach is evaluated on the following aspects which are the limitations identified in the existing researches:

*1) True parallel behavior in construction*
*2) Redundancy of the system*
*3) outcome of the system*

### A. Parallel Behavior of Robots

The approach developed has shown true parallel construction with multiple robots of the swarm simultaneously constructing the shape, placing building blocks. This is not observed in existing research which constructs given shapes [13], [15] or self-organizes into given shapes [16], [17]. Simple methods like blind bulldozing [9] shows true parallelism but with very limited decision making capabilities which does not suit for complex construction tasks such as the tasks we aim for in this research.

### B. System Redundancy

Usage of stigmergy has helped make the robots in our proposed system independent of each other. This makes the process of construction independent of the number of robots working. Only the speed of the construction is affected by the number of robots. In other existing methods, there are many methods with redundant swarms in both construction and self-organizing. However those do not show parallel construction behavior. Some do not have redundancy such as Werfel et al [13] method since it depends highly on the precompile instructions for each robot in the swarm.

### C. Output of the System

Our findings has shown that it can deliver an outcome as expected. It can construct a given 2D shape with multi robot collaboration. Similar researches that adapt swarm behavior for construction are successful in rule based shape generation [1], [12]. However their problem is the outcome not being pre-determined. Simple methods like blind bulldozing [9] are highly dependent on the environment and cannot be used for creating arbitrary shapes. The others which satisfies the final outcome, specially self-organizing in kilobot swarm [16], they do not display at least one property we consider in this comparison.

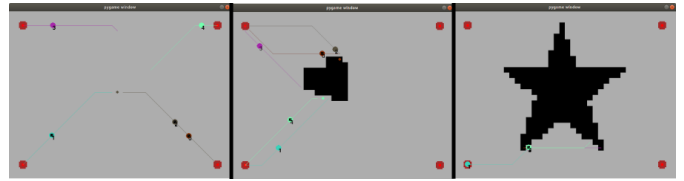

Fig. 3. The simulation showing start, an intermediate step and the end of construction of a pentagram shape. Four red locations near the four corners are building block factories. Colored circles and lines are swarm bots and their planned paths respectively.

## V. CONCLUSION

In this research we identified the limits of existing methods that are proposed for swarm construction tasks. Our proposed method was aimed at mitigating these limits identified resulting in a better and optimized swarm robotic system.

We simulated a swarm robotic system which does 2D construction tasks. We have used concepts of stigmergy and rule based control for the robots in the swarm and have developed a control algorithm for the robots to perform their individual tasks so that collectively the given shape is constructed. Our solution has shown to overcome the identified limitations of existing approaches.

Finally we can conclude that by introducing stigmergy with rule based control to a swarm robotic system with a carefully designed control algorithm can result in an optimized robotic swarm for construction tasks.

## VI. Future Works

The research has focused on 2D constructions as a beginner step to the real world 3D constructions. Furthermore, the shapes that are tested for construction are enclosed and did not have open areas in the middle (e.g. doughnut shaped or similar) or multiple separated construction areas

For real world applications, 3D construction optimized robots are needed in most cases. And complex shapes need to be built. Therefore the research needs to be extended into that area. Novel concepts can be introduced into the control algorithm in order to improve the optimization. Introduction of more swarm concepts into the system might help improve the system as well.